\documentclass[runningheads]{llncs}
\usepackage[T1]{fontenc}
\usepackage{graphicx}
\usepackage{hyperref}
\usepackage{color}

\usepackage{booktabs}
\usepackage{multirow}
\begin{document}
\title{Efficient Single Object Detection on Image Patches with Early Exit Enhanced High-Precision CNNs}
\titlerunning{Early Exit Enhanced High-Precision CNNs}
%
\author{Arne Moos\orcidID{0000-0003-2682-4422}}
\authorrunning{A. Moos}
%
\institute{TU Dortmund University, Robotics Research Institute\\ Otto-Hahn-Str. 8, 44227 Dortmund, Germany\\ \email{arne.moos@tu-dortmund.de}}
\maketitle              
\begin{abstract}
This paper proposes a novel approach for detecting objects using mobile robots in the context of the RoboCup Standard Platform League, with a primary focus on detecting the ball. The challenge lies in detecting a dynamic object in varying lighting conditions and blurred images caused by fast movements. To address this challenge, the paper presents a convolutional neural network architecture designed specifically for computationally constrained robotic platforms. The proposed CNN is trained to achieve high precision classification of single objects in image patches and to determine their precise spatial positions. The paper further integrates Early Exits into the existing high-precision CNN architecture to reduce the computational cost of easily rejectable cases in the background class. The training process involves a composite loss function based on confidence and positional losses with dynamic weighting and data augmentation. The proposed approach achieves a precision of 100\% on the validation dataset and a recall of almost 87\%, while maintaining an execution time of around 170 µs per hypotheses. By combining the proposed approach with an Early Exit, a runtime optimization of more than 28\%, on average, can be achieved compared to the original CNN. Overall, this paper provides an efficient solution for an enhanced detection of objects, especially the ball, in computationally constrained robotic platforms.

\end{abstract}
\section{Introduction}
\label{sec:Introduction}
Mobile robots require robust, reliable, and precise object detection capabilities to effectively perform their tasks. This paper focuses on the RoboCup Standard Platform League, which involves playing soccer using the NAO V6 humanoid robot platform\footnote{\url{https://www.aldebaran.com/en/nao}}. The robots use its cameras to detect objects in its environment, including static and dynamic ones like a rolling ball or other robots. Detecting dynamic objects can be challenging due to varying lighting conditions and fast movements. Deep neural networks with many layers are typically used, which increases the demand for computing power, a requirement that is lacking on a mobile robot platform like the NAO V6. Precise object detection is therefore more important than recall, as it is better to miss an object for a few frames than to have false detections and focus on the wrong areas. High precision also allows for faster re-detection of an object after it is lost, because it can be relied upon a single detection.

In robot soccer, the ball is the most critical object to detect because a match cannot be won without accurate detection of the ball. Detecting a rolling ball is crucial for the robot to react quickly. Therefore, its detection was studied in this paper. Conventional preprocessing techniques, such as scan lines, are used to identify a larger number of candidate regions where a ball may be present. However, these regions must be classified with a high precision. At the same time, the exact position of the ball, i.e. its center, within this patch must be determined, since it cannot be assumed that the candidate regions are always exactly centered on the object. Typically, the candidate regions' input data passes through a fixed neural network architecture. However, this fixed feed forward execution does not take into account that many of the patches that belong to the background class are more easily detectable and can therefore be rejected at earlier stages in the neural network.

This paper's main contribution consists of two parts. First, it presents a convolutional neural network architecture designed for computationally constrained robotic platforms, which is trained to achieve high precision classification of single objects in image patches and to determine their precise spatial positions. Second, the paper integrates Early Exits into an existing high-precision CNN architecture to reduce the computational cost of easily rejectable cases in the background class.

The remainder of this paper is organized as follows: Section~\ref{sec:RelatedWork} presents object detection techniques for resource-constrained robots, particularly in the context of the RoboCup Standard Platform League. Furthermore, related approaches concerning the use of Early Exits are discussed. Section~\ref{sec:Approach} explains the approach presented in this paper, including model design decisions, specialized training, and the addition of an Early Exit. The performance of the proposed approach is then evaluated in Section~\ref{sec:Evaluation}. Finally, Section~\ref{sec:Conclusion} concludes with a summary and an outlook.

\section{Related Work}
\label{sec:RelatedWork}
This section covers two different topics of related work. In the first subsection, we present the different ball detection algorithms used by several teams in the RoboCup Standard Platform League, which include the use of neural networks and specialized algorithms. In the second subsection, we will highlight the concept of Early Exit neural networks and the various techniques proposed by researchers to incorporate them into deep neural networks.

\subsection{Ball Detection in the RoboCup Standard Platform League}
In recent years, the RoboCup Standard Platform League has seen significant advances in ball detection algorithms for the NAO robot, especially since the transition to a black and white ball in 2016. These improvements have enabled robots to better detect, track, and respond to the ball during gameplay, resulting in more accurate and efficient play combined with passes.

Using a multistep process for ball detection, the B-Human team \cite{B-Human2019} scans for ball candidates using scan lines, followed by a neural network-based classification process to identify the real ball and estimate its center and radius. Their system includes three neural networks, one CNN for feature extraction and two DNNs for ball classification and position estimation. Similarly, the HTWK Robots team \cite{HTWK2020} uses a two-phase ball detection algorithm that involves an integral image and a deep convolutional neural network for hypothesis generation and classification, respectively. The rUNSWift team \cite{rUNSWift2019} uses a new convolutional neural network to improve their ball detection recall. Their framwork's ball candidate finder undergoes pre-processing, heuristic checks, and quality modifiers for consistent region of interest scaling. Using a candidate generator based on filtered segments and multiple neural networks, the HULKs team's \cite{HULKs2021} approach involves a pre-classification network for higher recall and a second classification network for higher precision. The position and radius of the ball are determined by a third neural network, which is optimized for maximum candidate throughput using a genetic algorithm. The Dutch Nao team \cite{DNT2018} developed an improved ball detection system, which uses a convolutional neural network for candidate generation and a field border detection system to reduce false positives. The Berlin United team \cite{Mellmann2019} proposes a two-step approach to detecting the ball used in competitions. Their approach involves finding candidates through perspective key points detection and classifying them using a measure function based on integral images. They also employ heuristics and neural networks to make the process tractable. 

Menashe et al. \cite{Menashe2018}, affiliated with the UT Austin Villa team, present an approach that combines color and texture features to distinguish the ball from the field and other objects in the image. The authors use a sliding window technique to localize the ball and apply a machine learning classifier to verify the detection. In \cite{Yan2019}, Yan et al. propose a real-time lightweight CNN for ball detection in robots with limited computational resources, utilizing a combination of convolutional and pooling layers to achieve high accuracy while keeping the model small. Additionally, the paper by O’Keeffe and Villing \cite{OKeeffe2018} proposes a benchmark data set and evaluation of deep learning architectures for ball detection in the RoboCup SPL, which can be used to compare the effectiveness of various ball detection approaches.

\subsection{Early Exit Neural Networks}
Deep neural networks (DNNs) have shown remarkable performance in various fields, such as computer vision and natural language processing. However, they are computationally expensive and require significant resources, hindering their deployment on resource-constrained devices. One approach to address this challenge is the use of Early Exits in DNNs. Early Exits allow a neural network to terminate its inference process early, bypassing unnecessary computations for some inputs, thereby reducing the overall computational cost.

Researchers have proposed methods for incorporating Early Exits in DNNs for efficient inference. Teerapittayanon et al. \cite{Teerapittayanon2016} introduced BranchyNet, a framework for fast inference via early exiting from DNNs by training auxiliary classifiers for intermediate layers. Huang et al. \cite{Huang2018} presented Multi-Scale Dense Networks, which utilize a dense connectivity pattern and multiple paths with different resolutions to enable Erly Exits. Figurnov et al. \cite{Figurnov2017} introduced Spatially Adaptive Computation Time for Residual Networks, which dynamically adjusts the computation time for different regions of an input image. Bolukbasi et al. \cite{Bolukbasi2017} proposed Adaptive Neural Networks for Efficient Inference, which use a reinforcement learning-based approach to decide when to exit early. Panda et al. \cite{Panda2016} presented Conditional Deep Learning, which employs an energy-based gating mechanism to selectively execute layers. Jayakodi et al. \cite{Jayakodi2018} proposed a co-design approach for trading-off accuracy and energy of deep inference on embedded systems. Berestizshevsky and Even \cite{Berestizshevsky2019} introduced cascaded inference based on soft max confidence, which dynamically sacrifices accuracy for reduced computation. Passalis et al. \cite{Passalis2020} proposed a hierarchical Early Exit approach that adapts the number and position of Early Exits for different input instances. Matsubara et al. \cite{Matsubara2022} identified the benefits of early exiting in split computing architectures, including reduced memory consumption, faster inference, and better load balancing across different processing units.

The approaches mentioned above have their unique strategies for Early Exits, and they can be categorized based on their input-adaptive, spatially adaptive, or hierarchical architectures. Input-adaptive methods dynamically adjust the network depth and width based on the input data \cite{Bolukbasi2017,Panda2016}. Spatially adaptive methods adjust the computation time or number of operations needed for different input regions \cite{Figurnov2017}. Hierarchical Early Exit methods split the computation into multiple parts and terminate the computation based on an Early Exit criterion \cite{Teerapittayanon2016,Huang2018,Passalis2020}. Other approaches aim to trade-off accuracy and energy by co-designing hardware and software for deep inference on embedded systems \cite{Jayakodi2018} or utilize cascaded inference based on soft max confidence, which dynamically sacrifices accuracy for reduced computation in \cite{Berestizshevsky2019}.

The input-adaptive and spatially adaptive methods are particularly useful for handling large variations in input data, such as in image classification tasks with varying sizes or aspect ratios. On the other hand, hierarchical Early Exit methods are more suitable for tasks with a clear hierarchy of features, such as in object detection or segmentation tasks. This paper falls into the latter category. However, the Early Exits are utilized quite differently from those mentioned before. In our case, there are only very few examples of the positive class to be recognized, but quite a few cases of the background class. Therefore, this paper presents an approach for accelerating the classification of the background class through an Early Exit enhancement.

\section{Approach}
\label{sec:Approach}
As described in Section~\ref{sec:Introduction}, this work's approach is to classify image regions (patches) obtained through preprocessing. Thereby, this work focuses on detecting the ball, which is one of the most crucial objects in robot soccer. In a worst-case scenario, where no ball is present in the image, the preprocessing stage generates up to 80 hypotheses per frame, with a mean and standard deviation of $30\pm10$ hypotheses.

Now, on the one hand, this leads to the constraint that the detection must be executed with a high precision of at least 99.99\%. Consequently, since in robot soccer there is only one ball on the field at a time, it also means that once a ball was detected with a high precision, processing all subsequent patches can be avoided for this frame. Nevertheless, it is imperative that the total execution time does not exceed the robot's real-time data processing capability, which is typically 30 FPS for the cameras. Thus, it is crucial to ensure that the robot can execute other important modules subsequently to ball detection.

Section~\ref{sec:Model} presents the neural network architecture we propose together with the chosen design decisions. Then, in Section~\ref{sec:Dataset} the dataset is explained, followed by a discussion on the training process in Section~\ref{sec:Training}. Finally, Section~\ref{sec:EarlyExit} presents the enhancement of the neural network using Early Exits.

\subsection{Model Architecture}
\label{sec:Model}
When designing a convolutional neural network (CNN) for detecting a ball in an image patch, the first and most important constraint considered is the execution time. For the framework running on the NAO V6, the ball detection should not exceed 8 ms in the normal case to provide enough buffer for the subsequent modules. When considering an average of 30 + 10 hypotheses per frame, this results in a maximum inference time of 0.2 ms per hypothesis.

Considering that convolution layers consume most of the execution time, it is apparent that there is room for optimization. Hence, we follow the MobileNet\cite{Howard2017,Sandler2018} approach, where computationally intensive convolutions are substituted with depthwise separable convolutions. A depth multiplier greater than 1 is utilized, temporarily increasing the number of filters for the depthwise convolution and subsequently reducing them for the pointwise convolution. This allows the extraction of more complex features while adhering to the execution time limitations.

Since most processors support SIMD instructions of some kind, including the NAO V6 with up to SSE 4.2, this possibility of parallel processing is also taken into account. With the NAO V6, four data types with 4 bytes (e.g., floats) can be processed simultaneously by SIMD instructions using a 128-bit register. To take advantage of this, care was taken in the design to ensure that the number of filters must be divisible by four.

The final model architecture for the CNN can be seen in Table~\ref{table:CNN}.

\begin{table}[t]
    \caption{Architecture of the CNN for the ball detection on an image patch. Each Separable-/Convolutional layer follows a Batch Normalization and a Leaky ReLu layer. The {*} marks the layer after which the Early Exit is attached.}
    \hfill{}%
    \begin{tabular}{ccccccc}
    Layer (type) & Filter & Kernel & Stride & Depth M. & \#MAC & Output\tabularnewline
    \toprule 
    Input & - & - & - & - & - & 32x32x3\tabularnewline
    \midrule 
    SeparableConv2D{*} & 8 & 3x3 & 2x2 & 1 & 13055 & 16x16x8\tabularnewline
    \midrule 
    Conv2D & 4 & 1x1 & 1x1 & - & 8190 & 16x16x4\tabularnewline
    \midrule 
    SeparableConv2D & 16 & 3x3 & 2x2 & 2 & 12800 & 8x8x16\tabularnewline
    \midrule 
    Conv2D & 8 & 1x1 & 1x1 & - & 8190 & 8x8x8\tabularnewline
    \midrule 
    SeparableConv2D & 20 & 3x3 & 2x2 & 4 & 14850 & 4x4x20\tabularnewline
    \midrule 
    Conv2D & 12 & 1x1 & 1x1 & - & 3840 & 4x4x12\tabularnewline
    \midrule 
    SeparableConv2D & 32 & 3x3 & 2x2 & 8 & 15745 & 2x2x32\tabularnewline
    \midrule 
    Conv2D & 16 & 1x1 & 1x1 & - & 2050 & 2x2x16\tabularnewline
    \midrule 
    Flatten & - & - & - & - & - & 64\tabularnewline
    \midrule 
    Dense & - & - & - & - & 192 & 3\tabularnewline
    \toprule  
    \multicolumn{7}{c}{Total \#MAC: 78912}\tabularnewline
    \multicolumn{7}{c}{Total \#Params: 6686}\tabularnewline
    \bottomrule
    \end{tabular}\hfill{}
    \label{table:CNN}
\end{table}

\subsection{Dataset}
\label{sec:Dataset}
The dataset we use for this work consists of small image patches in the RGB format with a size of 32x32 pixels. These were obtained from the preprocessing of the Nao Devils framework during games in recent years. A total of 225350 patches were labeled by hand. The dataset was then split in a ratio of around 70/30 between training and validation data. A detailed distribution can be found in Table~\ref{table:Dataset}. In addition to the initial classification, the following properties were also labeled:
\begin{itemize}
    \item \textbf{Bounding Box:} The upper left and lower right corners of the surrounding bounding box. Here, the coordinates can also be outside the patch because in the end, only the center of the object is used, so truncated objects can also be detected properly.
    \item \textbf{Concealed:} Indication of whether another object (i.e., in the foreground) partially conceals the object to be classified.
    \item \textbf{Visibility:} The visibility of the object in discrete increments with 25\% steps (i.e., 0-25\%, 25-50\%, \dots) based on the size of the bounding box that is inside the image, as well as the degree of concealment.
\end{itemize}
These additional properties account for image patch complexity in detection. A clear, fully visible ball being undetected is worse than a blurry or partially obscured ball.

\begin{table}[t]
    \caption{Number of patches in the dataset belonging to each class and its distribution
    to training and validation sets.}
    \hfill{}%
    \begin{tabular*}{8cm}{@{\extracolsep{\fill}}lccr}
    \toprule 
     & Ball & No ball & Total\tabularnewline
    \midrule
    Training & 69544 \textit{(44.07\%)} & 88274 \textit{(55.93\%)} & 157818\tabularnewline
    Validation & 28985 \textit{(42.92\%)} & 38547 \textit{(57.08\%)} & 67532\tabularnewline
    \bottomrule
    \end{tabular*}\hfill{}
    \label{table:Dataset}
\end{table}

\subsection{Training}
\label{sec:Training}
The training is conducted in TensorFlow\footnote{\url{https://www.tensorflow.org/}}, a software framework for machine learning. For the inference of the trained neural network on the NAO V6, TensorFlow Lite is used, which is specially designed for inference on mobile edge devices.
To enable the optimizer to perform effectively, a loss function that meets the requirements of the problem is needed. Since there are different objectives in the detection process, we present a composite loss function based on the two main objectives, combined with a dynamic weighting: 

\begin{itemize}
    \item \textbf{Confidence Loss:} Since the prediction of the confidence corresponds to probability distributions in the value range between 0 and 1, the use of a binary cross entropy seems to be appropriate. However, this loss function does not include any weighting to focus more on difficult examples. Therefore, the use of the Focal Loss \cite{Lin2020} is proposed, which is based on cross entropy but adds a weighting factor to down weight the nearly correct classified examples and thus focus more on difficult examples.

    \item \textbf{Positional Loss:} To evaluate the deviation of the position between ground truth and prediction, the Manhattan Distance is suggested. By using it, it is possible to determine the pixel difference between the true position and the prediction.

    \item \textbf{Dynamic Weighting:} The Dynamic Weighting has two parts. The first part uses dataset properties, as described in Section~\ref{sec:Dataset}, to penalize misclassification of simple examples more severely. This prioritizes objects that need to be recognized and increases recall of simple examples. The second part optimizes the training process for high precision by assigning each patch to one of four sections of the confusion matrix and multiplying them by weighting factors. To teach the neural network to avoid false positives, a large factor of $w_{fp}=1000$ is proposed.
\end{itemize}
At the end, the two loss functions are combined with specific weighting factors. The factors $w_c=1.0$ for the Confidence Loss and $w_p=0.5$ for the Positional Loss have proven to be effective in creating a total loss.

To improve the generalization of the neural network, data augmentation is employed. The amount of augmentation is dynamically controlled and gradually increased. Initially, affine transformations like scaling, translation, rotation, and shear are applied, as well as left-right flipping. Later, more augmentations such as brightness, contrast, and color changes, as well as motion blur and JPEG compression artifacts, are added.

\subsection{Adding an Early Exit}
\label{sec:EarlyExit}
As stated in Section~\ref{sec:Approach}, numerous ball hypotheses require classification. Not all images are of equal difficulty for classification. Therefore, it would be advantageous if the neural network is executed only up to the layer where precise prediction is possible to conserve computational time. Some methods for achieving this have already been introduced in Section~\ref{sec:RelatedWork}.

However, the distribution of object-to-background classes presented in this paper exhibits a substantial class imbalance, given that at most one ball should be present on the field/image. In our case, the primary goal is the rapid rejection of the background class, which only results in a change in recall, but it does not affect precision, which remains at a very high level.

Therefore, this paper proposes a method to enhance a deep neural network by adding an Early Exit to stop further inference when the background class has already been detected. This approach is generic and not specifically limited to the model presented in Section~\ref{sec:Model}. The procedure for inserting the Early Exit is as follows:

\begin{enumerate}
    \item Design a neural network model that satisfies the more precision targeted requirements. The execution time can be at the upper bound of the runtime limit.
    \item Train the model normally until there is no further improvement on the validation dataset. After training, lock all layer weights. In TensorFlow, this can be done using the \texttt{trainable} flag of the layers.
    \item Examine the neural network model, and we suggest inserting an Early Exit after the first convolutional layer. The combination of Max Pooling and Dense Layer has been found to be the most promising. Table~\ref{table:CNN} shows, marked with an *, after which layer the Early Exit is inserted for the CNN presented in this paper, while Table~\ref{table:EarlyExit} shows its structure. The Early Exit enhancement can be applied directly to the trained and locked model or a new model with transferred weights can be created.
    \item Train the layers of the Early Exit using the Confidence Loss, as described in Section~\ref{sec:Training}, with a high weighting factor for the false negatives $w_{fn}=100$. Achieving high recall is crucial for the Early Exit to avoid discarding potential positive objects too early.
    \item Separate the neural network at the Early Exit, resulting in two models. The first model uses the image patch as input and outputs the convolution output and the Early Exit classification. If the confidence at the Early Exit is high enough, indicating that the patch probably contains the expected object, execute the second model with the convolution output of the first model as its input.
\end{enumerate}

\begin{table}[t]
    \caption{Architecture of the Early Exit for the proposed ball detection CNN.}
    \hfill{}%
    \begin{tabular}{ccccc}
    Layer (type) & Pool Size & Stride & \#MAC & Output\tabularnewline
    \toprule 
    Input & - & - & - & 16x16x8\tabularnewline
    \midrule 
    MaxPooling2D & 2x2 & 2x2 & 510 & 8x8x8\tabularnewline
    \midrule 
    Flatten & - & - & - & 512\tabularnewline
    \midrule 
    Dense & - & - & 1025 & 1\tabularnewline
    \toprule  
    \multicolumn{5}{c}{Total \# MAC: 1535 \textit{(+1.95\%)}}\tabularnewline
    \multicolumn{5}{c}{Total \# Params: 513 \textit{(+7.67\%)}}\tabularnewline
    \bottomrule
    \end{tabular}\hfill{}
    \label{table:EarlyExit}
\end{table}

\section{Evaluation}
\label{sec:Evaluation}
For the evaluation of the developed CNN presented in Section~\ref{sec:Model}, the dataset described in Section~\ref{sec:Dataset} is utilized. As outlined in Section~\ref{sec:Training}, the ball detection CNN is initially trained without modification, after which it is enhanced with an Early Exit following the first convolutional layer and called EE-CNN. The evaluation criteria comprise both the runtime, as can be seen in Section~\ref{subsec:RuntimeComparison} and the performance, which is assessed using a confusion matrix with precision and recall subsequently determined shown in Section~\ref{subsec:Results}.

\subsection{Runtime Evaluation}
\label{subsec:RuntimeComparison}
In order to measure the runtime on the NAO robot, we use the TensorFlow lite runtime environment. In this process, 3600 measurements were performed, and the results are shown in Table~\ref{tab:ExecutionTimes}. 
It is directly evident that the runtime of the fully executed EE-CNN with 180 µs is 7.14\% slower than the original CNN. Because, as can be seen in Table~\ref{table:EarlyExit}, several new layers have been added that require additional computations. However, it can also be seen that the execution time up to the Early Exit with 64 µs needs about 62\% less runtime, which enables the approach presented in this paper to gain a performance advantage and the possibility to reduce the execution time. 

\begin{table}[tb]
\caption{Execution times measured on the NAO V6 over 3600 measurements.}
\label{tab:ExecutionTimes}
\hfill{}%
\begin{tabular*}{10cm}{@{\extracolsep{\fill}}lcccc}
\toprule 
 & Mean {[}ms{]} & Std {[}ms{]} & Min {[}ms{]} & Max {[}ms{]}\tabularnewline
\midrule
Full CNN & 0.168 & 0.077 & 0.129 & 1.411\tabularnewline
\midrule
\midrule 
\multirow{2}{*}{Full EE-CNN} & 0.180 & 0.083 & 0.136 & 1.406\tabularnewline
 & \textit{+7.14\%} & \textit{+7.79\%} & \textit{+5.43\%} & \textit{-0.36\%}\tabularnewline
\midrule
Early Exit & 0.064 & 0.049 & 0.043 & 1.299\tabularnewline
\bottomrule
\end{tabular*}\hfill{}
\end{table}

\subsection{Dataset Evaluation}
\label{subsec:Results}
To compare the performance of the new EE ball detection CNN to the original CNN, we executed both on the same training and validation dataset. Based on the predicted classifications, we calculated the confusion matrix and determined the precision and recall. Results on training data are provided as additional information only. The evaluation is performed solely on validation data. 
As can be seen in Table~\ref{tab:originalCNN}, the original CNN achieves a precision of 100\% on the validation dataset, with a recall of almost 87\%. This shows that the CNN presented here is able to detect many balls with a very high precision. Also, an average deviation of the ball center with around 0.471$\pm$0.795 pixel proves the effectiveness of the presented CNN model for detecting the correct spatial position.

When considering the Early Exit extended CNN in Table~\ref{tab:EE-CNN}, there is only a small change in recall with no change in precision. The latter is also clear, since the Early Exit presented in this work never contributes to a preliminary classification of the positive class, i.e., the ball. Thus, only the recall decreases slightly by 0.35\%.
On the other hand, the much more relevant part is shown in the last column, showing how often the Early Exit has decided to stop early. It can be seen that on the validation data set for roughly 43\% of the hypotheses, a decision can be made after the Early Exit without significantly influencing the recall. 

Based on the class distribution of the dataset shown in Table~\ref{table:Dataset}, the runtimes as shown in Table~\ref{tab:ExecutionTimes}, and the number of early exits, this leads to the shown mean execution time of 131 µs, which corresponds to a runtime optimization of more than 28\% compared to the original CNN.

\begin{table}[tb]
\caption{Results for the original CNN.}
\label{tab:originalCNN}
\hfill{}%
\begin{tabular*}{10cm}{@{\extracolsep{\fill}}lccccccc}
\toprule 
 & TP & FP & TN & FN & P & R & \phantom{EE}\tabularnewline
\midrule
Training & 59978 & 0 & 88359 & 9615 & 100\% & 86.18\% & \phantom{\textbf{\textit{41.85\%}}}\tabularnewline
\midrule
Validation & 25134 & 0 & 38581 & 3869 & 100\% & 86.66\% & \phantom{\textbf{\textit{42.57\%}}}\tabularnewline
\bottomrule
\end{tabular*}\hfill{}
\end{table}

\begin{table}[tb]
\caption{Results for the combined EE-CNN. The \#EE column specifies
how many times the Early Exit triggered in order to save computation time.}
\label{tab:EE-CNN}
\hfill{}%
\begin{tabular*}{10cm}{@{\extracolsep{\fill}}lccccccc}
\toprule 
 & TP & FP & TN & FN & P & R & \textbf{\#EE}\tabularnewline
\midrule
\multirow{2}{*}{Training} & 59769 & 0 & 88359 & 9824 & 100\% & 85.88\% & \textbf{66046}\tabularnewline
 & \textit{-0.35\%} & \textit{$\pm$0\%} & \textit{$\pm$0\%} & \textit{+2.17\%} & \textit{$\pm$0\%} & \textit{-0.3\%} & \textbf{\textit{41.85\%}} \tabularnewline
\midrule
\multirow{2}{*}{Validation} & 25033 & 0 & 38581 & 3970 & 100\% & 86.31\% & \textbf{28750}\tabularnewline
 & \textit{-0.40\%} & \textit{$\pm$0\%} & \textit{$\pm$0\%} & \textit{+2.61\%} & \textit{$\pm$0\%} & \textit{-0.35\%} & \textbf{\textit{42.57\%}} \tabularnewline
\midrule
\multicolumn{7}{r}{Mean Execution Time {[}\textmu s{]}} & \textbf{0.131}\tabularnewline
\multicolumn{7}{l}{} & \textbf{\textit{-28.24\%}}\tabularnewline
\end{tabular*}\hfill{}
\end{table}

\section{Conclusion and Future Work}
\label{sec:Conclusion}
The paper proposes a novel approach for object detection in mobile robots on computationally constrained platforms. The main focus is on detecting the ball in robot soccer games, where a high level of precision and real-time processing is required. The paper highlights the challenges in detecting dynamic objects in varying lighting conditions and fast movements, which requires a high level of computational power. The proposed approach can detect single objects in image patches and determine their precise spatial positions with a high precision classification. The proposed method utilizes a convolutional neural network with depthwise separable convolutions, which is optimized to achieve the highest possible accuracy while adhering to the time constraints. The paper also explores the concept of Early Exit neural networks and its potential for reducing computational costs while maintaining performance. Early Exits are integrated in order to terminate the network's inference process early, thereby reducing computational costs. This approach is evaluated and compared to the original CNN, which shows a decrease in the average execution time by 28\% for the Early Exit version, with equal precision and almost equal recall.

Future work could focus on optimizing the network architecture further to reduce the computational cost and increase the speed of execution. Additionally, exploring other methods for Early Exits and combining them with other techniques, such as pruning or quantization, could result in more efficient and accurate object detection in mobile robots. Finally, investigating the robustness of the proposed approach to changing lighting conditions and fast movements in various game scenarios as well as a different class distribution could further improve its applicability in practical use cases.

%
%
%
\bibliographystyle{splncs04}
\bibliography{literature}

\end{document}